\documentclass[runningheads]{llncs}
\usepackage[T1]{fontenc}
\usepackage{graphicx}
\usepackage{amsmath,amsfonts}
\usepackage{amssymb}
\usepackage{multirow}
\usepackage{enumerate}
\begin{document}
\title{Syntax Tree Constrained Graph Network for Visual Question Answering}
%
%
\author{Xiangrui Su\textsuperscript{\rm \textdagger} \inst{1}\and
Qi Zhang\textsuperscript{\rm \textdagger} \inst{2,3} \and Chongyang Shi\thanks{Corresponding author. \email{cy\_shi@bit.edu.cn}\\
\noindent \textsuperscript{\rm \textdagger} The first two authors contribute equally to this work.}\inst{1}\ \and
Jiachang Liu\inst{1}\and Liang Hu\inst{2,3}
}
%
\authorrunning{X. Su et al.}
%
\institute{
Beijing Institute of Technology, Beijing, China\\
\and
Tongji University, Shanghai, China\\
\and
DeepBlue Academy of Sciences, Shanghai, China
}
\maketitle              
\begin{abstract}
Visual Question Answering (VQA) aims to automatically answer natural language questions related to given image content. Existing VQA methods integrate vision modeling and language understanding to explore the deep semantics of the question. However, these methods ignore the significant syntax information of the question, which plays a vital role in understanding the essential semantics of the question and guiding the visual feature refinement. To fill the gap, we suggested a novel Syntax Tree Constrained Graph Network (STCGN) for VQA based on entity message passing and syntax tree. This model is able to extract a syntax tree from questions and obtain more precise syntax information. Specifically, we parse questions and obtain the question syntax tree using the Stanford syntax parsing tool. From the word level and phrase level, syntactic phrase features and question features are extracted using a hierarchical tree convolutional network. We then design a message-passing mechanism for phrase-aware visual entities and capture entity features according to a given visual context. Extensive experiments on VQA2.0 datasets demonstrate the superiority of our proposed model.

\end{abstract}
\section{Introduction}
Visual Question answering (VQA) aims to automatically answer natural language questions related to a given image content. It requires both computer vision technology to understand the visual content of images and natural language processing technology to understand the deep semantics of questions. VQA has various potential applications, including image retrieval, image captioning, and visual dialogue systems, therefore becoming an important research area.

Recently, various VQA methods~\cite{ref_ban,ref_murel,ref_lcgn,ref_regat} have been proposed to capture significant question semantics and visual features by mining explicit or implicit entity relationships. For example, BAN~\cite{ref_ban} considers the bilinear interaction between two sets of input channels of images and questions by calculating the bilinear distribution of attention to fuse visual and textual information. Murel et al.~\cite{ref_murel} design an atomic inference unit to enrich the interaction between the problem region and the image region and optimize the visual and problem interaction by using a unit sequence composed of multiple atomic units. LCGN~\cite{ref_lcgn} designs a question-aware messaging mechanism, uses question features to guide the refinement of entity features based on the entity complete graph, and realizes the integration of entity features and context information. However, these methods typically capture explicit or implicit entity relationships in images while ignoring the syntax relation between words, which contributes to capturing the deep semantics of the question.

\begin{figure}[!t]
     \centering
    \vspace{-2mm}
      \includegraphics[width=0.5\textwidth]{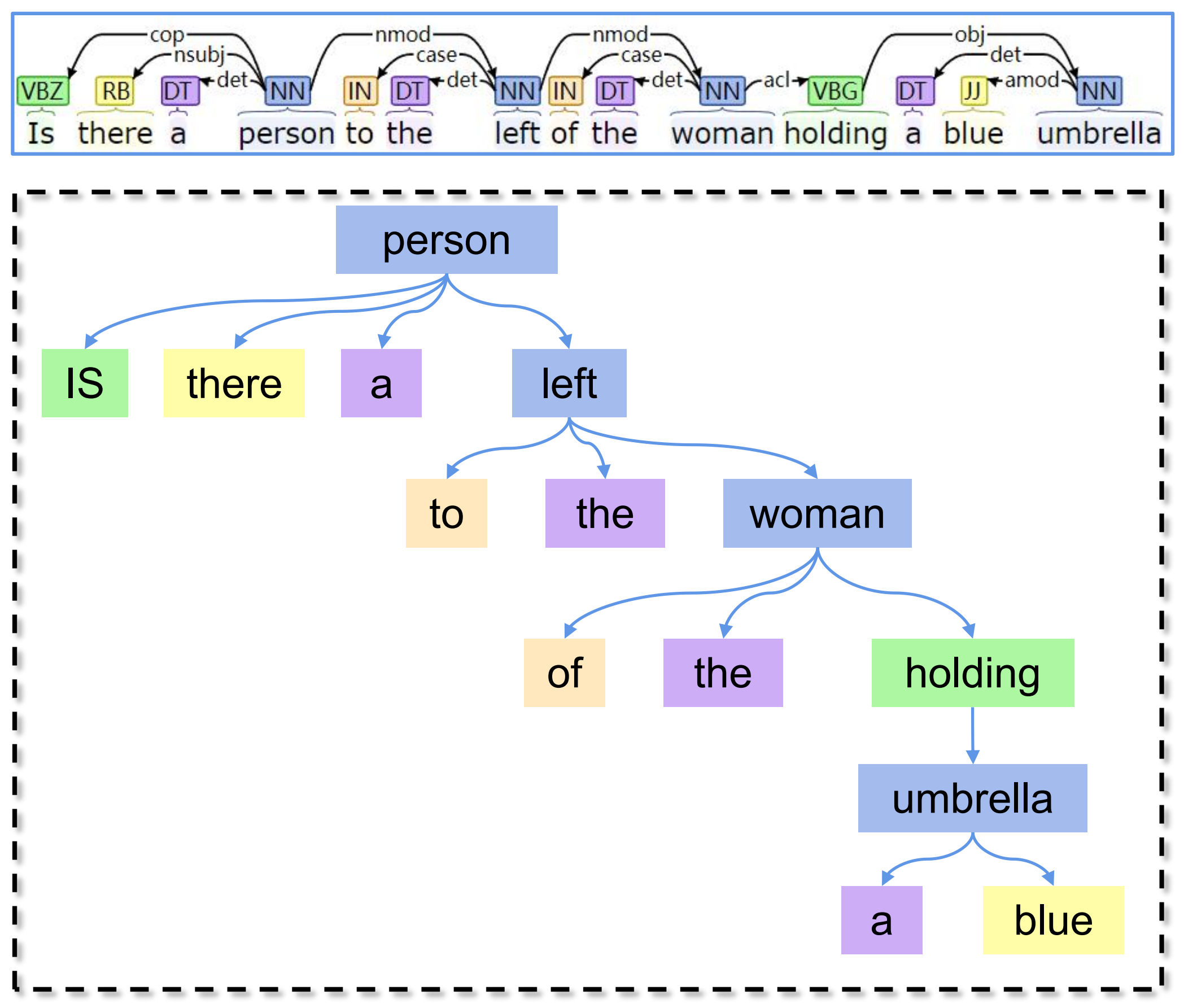}
      \includegraphics[width=0.46\textwidth]{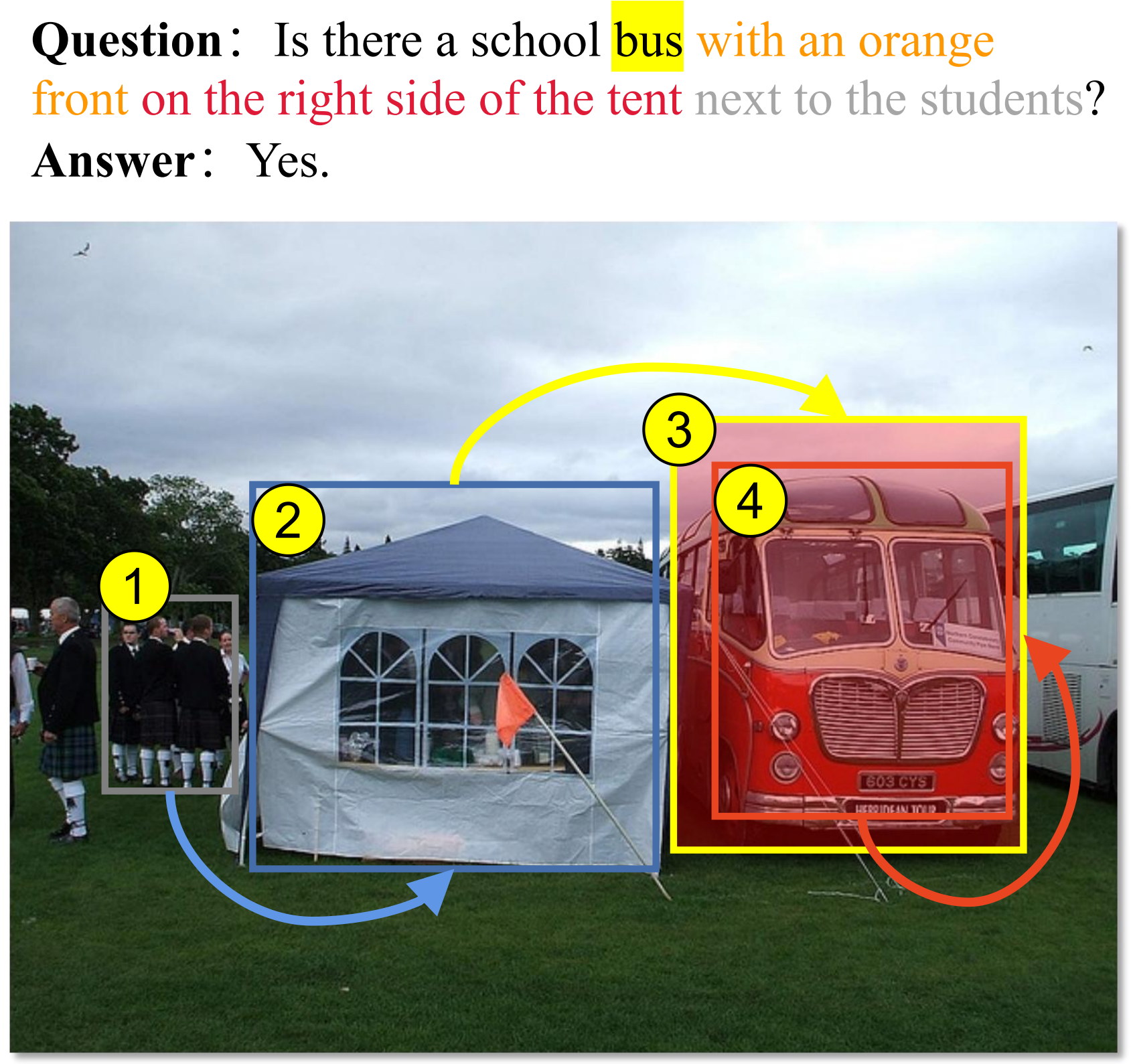}
      \vspace{-3mm}
    \caption{Examples of a syntax tree and message passing respectively derived from the VQA2.0 dataset. The left figure shows a syntax three, and the right figure shows the message passing mechanism.}
    \label{fig:example}
    \vspace{-6mm}
\end{figure}

Intuitively, using a syntax tree in VQA tasks has two major benefits. First, questions are usually short in length, and adding more syntactic information is necessary to understand the semantics of the questions. Secondly, the syntax tree hierarchically organizes keywords and context words through a tree structure, which is effective for capturing key information in the questions. As shown in the illustration in Fig. \ref{fig:example} (a), the words "person", "left", and "woman" which are far apart in the original question are adjacent in the syntax tree. These three words are the core information of this question. Therefore, the syntax tree can better capture the key information in terms of feature extraction.

Besides, in the field of VQA, images are the key information source to infer answers. It is also one core objective of the VQA model to understand the information in images. Since images are composed of many visual entities, there are many implicit relationships between these entities. Intuitively, it is necessary to perceive these implicit relationships and achieve effective message passing between entities for obtaining the entity features of scene context awareness.

As shown in Fig. \ref{fig:example} (b), in the process of answering the question, Entity 1 firstly transmits its own features to Entity 2 based on the phrase "next to the students"; Entity 2 and Entity 4 then pass on its features to Entity 3 based on the phrase "on the right side of the tent" and "with an orange front", and Entity 3 is accordingly able to integrate information from surrounding entities to answer the question more accurately. Obviously, phrase features can be used to guide entities to carry out targeted message passing, such that it is necessary for the VQA system to pay more attention to the area most relevant to the question.


In light of the above observation, we propose a Syntax Tree Constrained Graph Network (STCGN) by modeling the syntax tree and entity message-passing mechanism. The main contributions of this paper are summarized:
\begin{itemize}
    \item We propose a novel VQA model, named STCGN, which is equipped with a tree hierarchical convolutional network that learns syntax-aware question representation and phrase representation comprehensively, a phrase-aware entity message passing network that captures context-aware entity representations of visual scenes.
    \item Extensive experimental results on VQA2.0 datasets demonstrate the significant superiority of our STCGN and prove the validity of phrase-aware messaging network through visualization experiments.
\end{itemize}

\section{Related Work}
Recently, various methods have been proposed to improve the accuracy of visual question answering. The image is typically represented by a fixed-size grid feature extracted from a pre-trained model such as VGG~\cite{SimonyanZ14a} and AlexNet~\cite{00200CSWL22}, and the question is typically embedded with LSTM~\cite{ref_lstm}. Both of these features are combined by addition or element-wise multiplication~\cite{ref_simple,ZhangCSN22}. But using the base fusion method to achieve feature fusion is too simple to capture a key part of the image. Therefore some neural Network based method~\cite{ren2015exploring} is proposed to fuse both the visual and question features. For example, Ren~\cite{ren2015exploring} propose an LSTM-based fusion strategy by projecting extracted visual feature into the same word embedding space. However, not all of the pictures are strongly related to the question, thus some of the non-relevant grids in the picture should be filtered.

The attention mechanism~\cite{ref_butd,nguyen2018improved,ilievski2017generative,YeH0LNL23} is used to calculate the importance of each grid area. For example, Bottom-Up and Top-Down attention (BUTD)~\cite{ref_butd} uses a bottom-up and top-down approach to capture attention. In the bottom-up process, they used Faster-Rcnn to extract the features of all objects and their significant regions. In a top-down process, each attention candidate is weighted using task-specific context to obtain a final visual representation. Besides, the bilinear models~\cite{yu2018beyond,nguyen2018improved,ref_mutan,guo2021bilinear} show a strong ability of cross-modal feature interaction. For example, MFH~\cite{yu2018beyond} fully mines the correlation of multi-mode features through factorization and high-order pooling, effectively reducing the irrelevant features, and obtaining more effective multi-mode fusion features. BGN~\cite{guo2021bilinear} on the basis of the BAN~\cite{ref_ban} model designs a bilinear graph structure to model the context relationship between text words and visual entities and fully excavates the implicit relationship between text words and visual entities. However, these methods do not pay enough attention to the entity relationship in the pictures and the relationship between the words in the questions. Extracting entity features can also be optimized through implicit or explicit relational modeling.

To address the above problem, some relation-based VQA methods, e.g., Murel~\cite{ref_murel}, LCGN~\cite{ref_lcgn}, and ReGAT~\cite{ref_regat}, and more effective attention mechanisms, e.g., UFSCAN~\cite{UFSCAN} and MMMAN~\cite{MMMAN} have been proposed. Murel, LCGN, and ReGAT refine visual representations by explicitly or implicitly modeling relationships and enhance feature interactions between modalities. UFSCAN and MMMAN improve attention scores in visual areas through more efficient attention mechanisms. Compared to these prior works, Our model proposes a syntax-aware tree hierarchical convolutional network to extract syntax relation-aware question representation and phrase representation from questions. It further proposes a phrase-aware message passing network to capture the entity features of visual scene context-awareness based on implicit entity relations.


\section{The STCGN Model}
Given a question $q$ and a corresponding image $I$, the goal of VQA is to predict an answer $\hat{a} \in \mathcal{A}$ that best matched the ground-truth answer $a^*$. Following previous VQA models, it can be defined as a classification task:
\begin{align}
    \hat{a} = arg\max_{a\in \mathcal{A}}F_{\theta }(a|I,q)
\end{align}
where $F$ is the trained model, and $\theta$ is the model parameter.

Denote the given picture as $\mathcal{I}$, $K$ visual entities are extracted from it using Faster R-CNN, and the feature representation of the i-th entity is denoted as $\mathbf{v}_i$. Meanwhile, we have a picture-related question consisting of N words $\mathcal{Q}=(q_1,q_2,...,q_N)$. The features of each word are initialized using the Glove~\cite{ref_glove} word embedding model and the word sequence feature is $\mathbf{X}=(\mathbf{x}_1,\mathbf{x}_2,...,\mathbf{x}_N)$. The task of the visual question answering is to use image and text information, extract relevant features and fuse them to generate an answer probability distribution vector $\mathbf{\hat{p}}=(\hat{p}_1,\hat{p}_2,...,\hat{p}_{N_{ans}})$, where $N_{ans}$ denotes the number of categories of answers and $\hat{p}_i$ denotes the probability that the i-th answer is the final answer. We take the answer with the highest probability in $\mathbf{\hat{p}}$ as the final predicted answer of the visual question answering system.

\subsection{Network Architecture}
\begin{figure}
    \centering
    \vspace{-2mm}
    \includegraphics[width=0.9\textwidth]{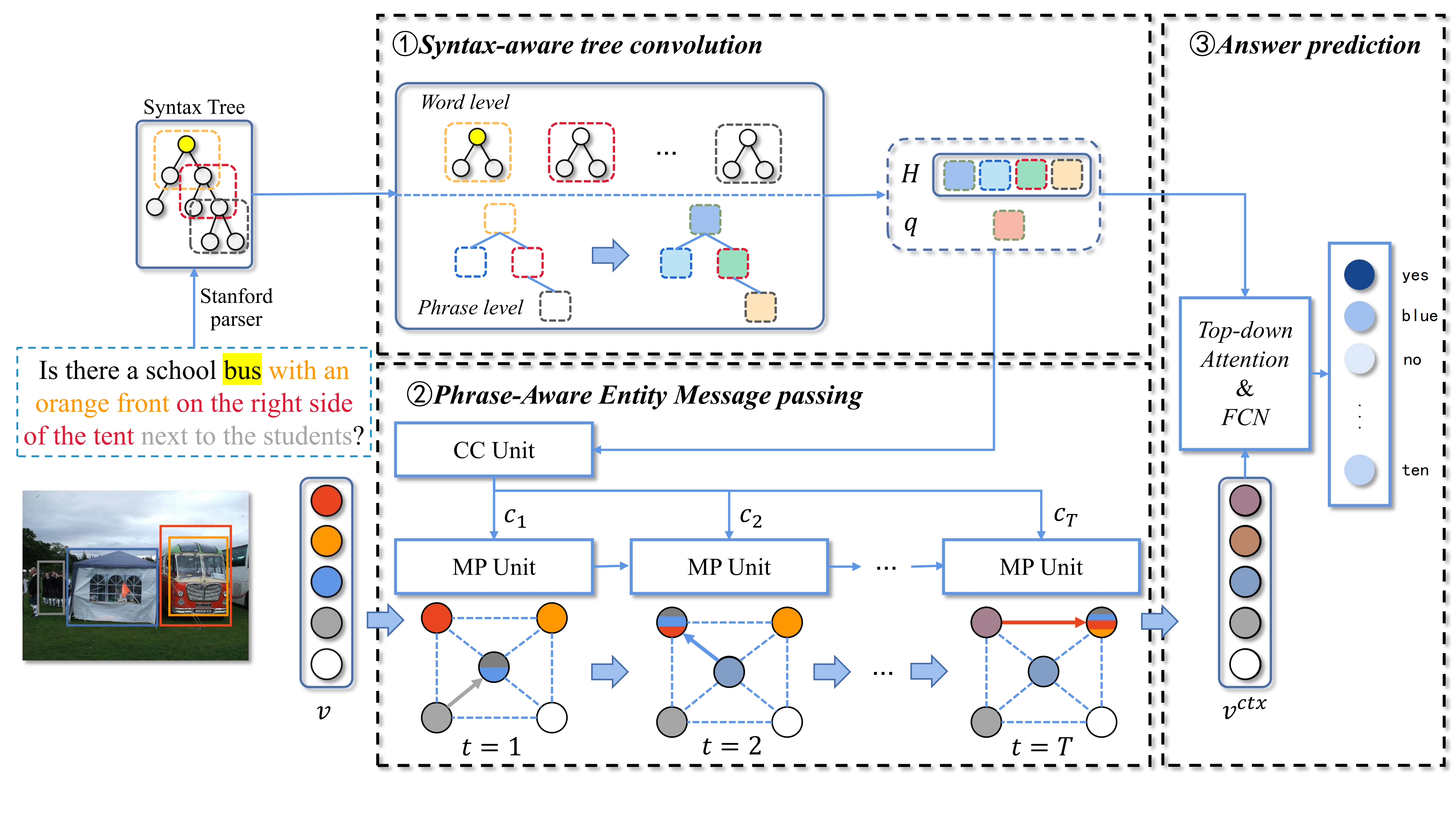}
    \vspace{-4mm}
    \caption{The network architecture of our STCGN model.}
    \label{fig:STCGN}
    \vspace{-5mm}
\end{figure}

The architecture of STCGN is shown in Fig. \ref{fig:STCGN}, which consists of three main modules: 
(1) \textbf{Syntax-aware Tree Convolution} module utilizes the syntax tree of the question and uses the tree hierarchical convolution model to extract the syntax-aware phrase features and question features.
(2) \textbf{Phrase-aware Entity Message Passing} module discovers the implicit connections between visual entities and relevant message features based on the syntax-aware phrase features, question features, and entity complete graphs, and then builds the scene context-aware visual entity features.
(3) \textbf{Top-down Attention-based Answer Prediction} module fuses the question features and visual entity features by using Top-down attention and performs the final answer prediction. 

\subsection{Syntax-aware Tree Convolution}
In VQA tasks, understanding the essence of the question correctly is the first priority to make a good answer. Syntax information plays a very important role in the question text because it can offer important help in understanding the question. The syntax information contains the dependencies between words and the part of speech (POS) of these words. Therefore, we can construct a syntax tree $\mathcal{T}=(\mathcal{Q}, \mathcal{E})$ based on the dependencies between words. By observing Fig. \ref{fig:example} (b), we can see that the phrases in question play an important role in instructing visual entities for message passing. Therefore, we propose a tree-based hierarchical convolutional network to model the syntax-aware phrase features. The network consists of two layers: word-level convolution and phrase-level convolution.

For word-level convolution, we first construct the syntax subtree with each non-leaf word node and its direct children in the syntax tree. In this way, we can decompose the syntax tree into a set of syntax subtree $F=(f_1,f_2,... ,f_s)$, where each subtree $f_i=(q_i,q_{c1},q_{c2},... ,q_{cn})$ as the convolution unit, $cn$ denotes the number of children of $i$ word nodes. Furthermore, we propose a convolution method based on text convolution~\cite{zhang2015sensitivity}. First, we use the Glove word embedding model to map $f_i$ to high-dimensional dense word features. Then, we obtain the learnable POS feature vector by using a POS feature dictionary of length 42 for random initialization. As a result, we can obtain the sequence of POS features in the question $\mathbf{T}_i = (\mathbf{t}_i,\mathbf{t}_{c1},... ,\mathbf{t}_{cn})$. Finally, we concatenate word features and POS features to obtain word-level convolutional input features $\mathbf{X}^{cat}_i$ of words:
\begin{gather}
    \mathbf{X}_i= \mathrm{Glove}(f_i),\ 
    \mathbf{T}_i = \mathrm{RandomInit}(t_i,t_{c1},...,t_{cn}),\ 
    \mathbf{X}^{cat}_i = \left [\mathbf{X}_i\oplus \mathbf{T}_i \right]
\end{gather}
We define a text convolution kernel $\mathbf{G}$ for each syntax subtree. First, we concatenate the input features $\mathbf{X}^{cat}_i$ to extract the key information in the text. Then, we use the maximum pooling for further feature filtering.
\begin{gather}
   \mathbf{g}_i = \max(\hat{\mathbf{g}}_i),\quad s.t.,\  \hat{\mathbf{g}}_i = \mathrm{ReLU}(\mathbf{G}*[\mathbf{X}^{cat}_i\oplus\mathbf{X}^{cat}_{c1}\oplus ...\oplus\mathbf{X}^{cat}_{cn}]+\mathbf{b}_i)
\end{gather}
$*$ indicates the text convolution, and $\mathbf{b}_i$ is the offset term.

For the phrase-level convolution, we design the syntax relation-aware graph attention network for capturing syntax phrase features $\mathbf{h}_i$ from multiple aspects based on the convolution features $g_i$ of each subtree and the syntax tree $\mathcal{T}$:
\begin{gather}
    \mathbf{h}_i^{*}=||^{M}_{m=1}\sigma \left(\sum_{j \in \mathcal{N}_i}\alpha_{ij} \cdot \mathbf{W}_{dir(i,j)}\mathbf{g}_j + \mathbf{b}_{dep(i,j)} \right)\\
    \alpha_{ij} = \frac{exp((\mathbf{U}\mathbf{x}_i^{'})^{\top } \cdot \mathbf{V}_{dir(i,j)}\mathbf{g}_j+\mathbf{b}^{'}_{dep(i,j)}  )}
{\sum_{j \in \mathcal{N}_i} exp((\mathbf{U}\mathbf{x}_i^{'})^{\top } \cdot \mathbf{V}_{dir(i,j)}\mathbf{g}_j+\mathbf{b}^{'}_{dep(i,j)}} 
\end{gather}
where $\mathbf{W}_{\{\cdot\}} \in \mathbb{R}^{(d_h/M) \times (d_x+d_t)}$, $\mathbf{V}_{\{\cdot\}}\in \mathbb{R}^{(d_h/M) \times (d_x+d_t)}$, and $\mathbf{U} \in \mathbb{R}^{(d_h/M) \times (d_x+d_t)}$ are parameters matrices, $\mathbf{b}_{\{\cdot\}}$, $\mathbf{b}_{\{\cdot\}}^{'}$ are offset terms, $dir(i,j)$ denotes the direction of each relation, and $dep(i,j)$ denotes the different kinds of dependencies.

Then, to capture the sequence correlation between phrase features and to obtain the features of the whole sentence, we performed phrase sequence feature extraction using a bidirectional GRU ($\operatorname{biGRU}$) network:
\begin{gather}
    h_t = \operatorname{biGRU}([h_{t-1},\mathbf{h}_i^{*}])
\end{gather}
where $\mathbf{W}_z \in \mathbb{R}^{d_h \times (2*d_h)}$, $\mathbf{W}_r \in \mathbb{R}^{d_h \times (2*d_h)}$, $\mathbf{W}_h \in \mathbb{R}^{d_h \times (2*d_h)}$ are the parameter matrices shared by the network, and $\sigma$ denotes the $sigmoid$ function.

The final output is $\mathbf{H}=(\mathbf{h}_1,\mathbf{h}_2,... ,\mathbf{h}_s) \in \mathbb{R}^{s \times (2*d_h)}$ with $\mathbf{q} \in \mathbb{R}^{2*d_h}$, where
$\mathbf{h}_{i}=[\overrightarrow{h}_i \oplus \overleftarrow{h}_i ]$, $\mathbf{H}$ represents the syntax-aware phrase feature sequence, $\mathbf{q}=\mathbf{h}_s$, and $\mathbf{q }$ is the output of the last hidden layer, which incorporates the information of all iteration steps and can represent the syntax-aware question features.

\subsection{Phrase-aware Entity Message Passing}
The inputs to the phrase-aware entity message passing module include the syntax-aware question feature $\mathbf{q}$, the syntax-aware phrase feature $\mathbf{H}=(\mathbf{h}_1,\mathbf{h}_2,... ,\mathbf{h}_s)$ and the original visual entity features $\mathbf{V}=(\mathbf{v}_1,\mathbf{v}_2,... ,\mathbf{v}_K)$. We propose a phrase-aware multi-step instruction calculation method, which sends each word separately with question features into the instruction calculation network at each time step to calculate the contribution of each word to the question. Then, we weight the words according to their contribution levels to obtain the instruction vector $\mathbf{c}_{t} \in \mathcal{R}^{d_c}$ that guides the visual entities for message passing:
\begin{gather}
    \mathbf{c}_{t}=\sum_{i=1}^{N} \alpha_{t, i} \cdot \mathbf{h}_{i},\quad \alpha_{t, i} =\underset{i}{softmax}\left(\mathbf{W}_{1}\left(\mathbf{h}_{i} \odot\left(\mathbf{W}_{2}^{(t)} \operatorname{ReLU}\left(\mathbf{W}_{3} \mathbf{q}\right)\right)\right)\right)
\end{gather}
where $\mathbf{W}_{1},\mathbf{W}_{3}$ are parameter matrices shared by all iteration steps, while $\mathbf{W}_{2}^{(t)}$ specifies to time step $t$ to generate different messages at different iteration steps.

We use $w_{j,i}^{(t)}$ to denote the weight of message delivered from $j$ entities to $i$ entities at time step $t$ and $m_{j,i}^{(t)}$ to denote the sum of messages received by the $i$th entity from other entities at time step $t$.
At time step $t$, taking the $i$-th entity as an example, we compute the features of messages delivered by adjacent entities to the $i$-th vector based on the instruction vector $\mathbf{c}_{t}$:
\begin{gather}
    \tilde{\mathbf{v}}_{i,t}=\left[\mathbf{v}_i\oplus \mathbf{v}_{i,t-1}^{ctx} \oplus ((\mathbf{W}_4\mathbf{v}_i)\odot(\mathbf{W}_5\mathbf{v}_{i,t-1}^{ctx}))\right] \\
    w_{j,i}^{(t)} = softmax\left((\mathbf{W}_6\tilde{\mathbf{v}}_{i,t})^{\top}(\mathbf{W}_7\tilde{\mathbf{v}}_{j,t})\odot(\mathbf{W}_8\mathbf{c}_t) \right)\\
    m_{j,i}^{(t)} = w_{j,i}^{(t)}\cdot((\mathbf{W}_9\tilde{\mathbf{v}}_{j,t})\odot(\mathbf{W}_{10}\mathbf{c}_t))
\end{gather}
And then we use the residual network for message aggregation to achieve the recognition of the scene context by the current visual entity $\mathbf{v}_{i,t}^{ctx}$. Finally, we concatenate the original entity features $\mathbf{v}_i$ with the scene context features $\mathbf{v}_{i,T}^{ctx}$ and obtain the final scene context-aware entity features $\mathbf{v}_{i}^{out}$:
\begin{equation}
    \mathbf{v}_{i}^{out} = \mathbf{W}_{12}\left[\mathbf{v}_{i}\oplus \mathbf{v}_{i,T}^{ctx} \right],\quad \mathbf{v}_{i,t}^{ctx} = \mathbf{W}_{11}\left[\mathbf{v}_{i,t-1}^{ctx}\oplus \sum_{j=1}^{K}m_{j,i}^{(t)} \right]
\end{equation}
where $\mathbf{W}_{\{\cdot\}}$ denotes the parameter matrix shared by all iteration steps.

\subsection{Top-down Attention-based Answer Prediction}
Given a picture $\mathcal{I}$, we combine the set of visual entity features $\{\mathbf{v}_i\}_{i=1}^K$ with syntax-aware question features $\mathbf{q}$ and use a top-down attention mechanism for feature fusion. This mechanism first calculates the attention scores between each visual entity feature and the question feature as follows:
\begin{equation}
    \beta_i =\underset{i}{softmax}\left(\mathbf{W}_{13}tanh(\mathbf{W}_{14}\mathbf{v}^{out}_i + \mathbf{W}_{15}\mathbf{q}))\right)
\end{equation}
where $\mathbf{W}_{13},\mathbf{W}_{14}, \mathbf{W}_{15}$ denote the weight matrices. Then, we weigh each visual entity feature using a top-down attention mechanism and perform a nonlinear transformation of the joint features by a two-layer perceptron model. Then, we calculate the probability score for each answer using the $softmax$ function:
\begin{equation}
    \mathbf{p} = \mathbf{W}_{16}ReLU\left(\mathbf{W}_{17} \left [\sum_{i=1}^{N}\beta_i\mathbf{v}_i^{out} \oplus \mathbf{q} \right] \right) 
\end{equation}
where $\mathbf{W}_{16},\mathbf{W}_{17}$ are the weight matrices, and $ReLU$ is the activation function. 

\subsection{Loss Function}
Finally, we train our model by minimizing the cross-entropy loss function:
\begin{equation}
    \mathcal{L} = - \frac{1}{N_{ans}}\sum_{i=1}^{N_{ans}}\left(y_i\cdot log(\hat{p}(y_i)) + (1 - y_i)\cdot log((1-\hat{p}_(y_i)))\right)
\end{equation}
where $N_{ans}$ denotes the number of answer categories, and $y_i$ is defined as $y_i=min(\frac{\#humans\ provided\ ans}{3},1)$ 
where $\#humans\ provided\ ans$ denotes the total times by which the answer was selected during the data annotation period. $\hat{p}(y_i)$ denotes the probability that the output belongs to the $i$-th class of answers.
\section{Experiments}

\subsection{Experimental Settings}
\textbf{Datasets}: We adopt two datasets: 1) MSCOCO~\cite{ref_coco} with more than 80k images and 444k questions (training), 40k images and 214k questions (test), 80k images, and 448k questions (validation), respectively. \textbf{Baselines}: We select the following $11$ state-of-the-art methods as baselines to evaluate the performance of our STCGN: LSTM-VGG~\cite{ref_simple}, SAN~\cite{ref_san}, DMN~\cite{ref_dmn}, MUTAN~\cite{ref_mutan}, BUTD~\cite{ref_butd}, BAN~\cite{ref_ban}, LCGN~\cite{ref_lcgn}, Murel~\cite{ref_murel}, ReGAT~\cite{ref_regat}, UFSCAN~\cite{UFSCAN}, MMMAN~\cite{MMMAN}. Refer to Appendix \ref{app:exp_settings} for more details about datasets, baselines, and settings.

\subsection{Performance Comparison}
Table 1 shows the overall performance of all comparative methods, with the best results highlighted in boldface, where we draw the following conclusions:

1) LSTM+VGG adopts the classic VQA framework with a single model module and only uses a simple vector outer product to achieve feature fusion. The modal information interaction is too simple, resulting in the performance of the final model being inferior to other models.
    
2) SAN and DMN adopt the typical NMN architecture to decompose the VQA task into a sequence of independent neural networks executing subtasks. The SAN achieves multi-step queries by stacking multiple attention modules, which significantly outperforms the LSTM+VGG network in terms of performance. DMN structure is more modular, using a dynamic memory network module to achieve episodic memory ability, all modules cooperate to complete the question-answering task, in the ability to answer three types of questions is better than the previous two models.

\begin{table}[ht]
    \centering
    \vspace{-5mm}
    \begin{tabular}{|c|c|c|c|c|c|c|c|c|}
        \hline
        {\multirow{2}{*}{\textbf{method}}} & \multicolumn{4}{c|}{\textbf{Test-std}} & \multicolumn{4}{c|}{\textbf{Test-dev}}\\
        \cline{2-9}
         & Overall & Y/N & Num & Other & Overall& Y/N & Num & Other  \\
        \hline
        LSTM+VGG  & 58.2 & 80.41 & 37.12 & 44.21  & 57.13 & 80.52 & 36.74 & 43.18\\
        SAN  & 58.9 &80.56 &36.12 &46.73  & 58.70 &79.3 & 36.6 & 46.10\\
        DMN  & 61.28 &81.64 &37.81 &48.94 & 60.30 & 80.50& 36.8 & 48.3\\
        MUTAN  &63.58 &80.35 &40.68 & 53.71 & 62.70 & 79.74 & 40.86 & 53.13\\
        BUTD  &65.42 &82.13 &44.25 &56.96 & 64.37 & 81.12 & 43.29 & 56.87\\
        BAN  &70.31 &86.21 &50.57 &60.51 & 69.11 & 85.24 & 50.03 & 60.32\\
        BAN + Counter  &70.89 &86.44 & 55.12 &60.67 & 70.02 & 85.44 & \textbf{54.21} & 60.25\\
        \hline
        MuRel  & 68.41 &84.22 &50.31 &58.49 & 68.03 & 84.77 & 49.84 & 57.75\\
        LCGN  &69.44 &85.23 &49.78 &59.37 &68.49 & 84.81 & 49.02 &58.47\\ 
        ReGAT  &70.54 &86.34 &54.14 &60.76 &70.12 &86.02 &53.12 & 60.31\\
        UFSCAN & 70.09 & 85.51 & \textbf{61.22} & 51.46 & 69.83 & 85.21 & 50.98 & 60.14\\
        MMMAN  &70.28 & \textbf{86.69}& 51.83& 60.22 & 70.03 & \textbf{86.32} & 52.21 & 60.16 \\
        STCGN  & \textbf{70.99} &85.77 &50.46 &\textbf{60.79} &\textbf{70.21} & 85.49 & 50.11 & \textbf{60.40}\\
        \hline
    \end{tabular}
    \vspace{-3mm}
    \caption{Performance comparison on the VQA2.0 dataset.}
    \label{tab:Comp_exp}
    \vspace{-6mm}
\end{table}

3) MUTAN, BUTD, BAN, and BAN+Counter are typical VQA models based on bilinear pooling and attentional mechanisms. Compared with classical frameworks and NMN architectures, these four models have more fine-grained feature extraction and feature fusion, significantly improving performance. In particular, the bilinear attention mechanism of BAN fully exploits the implicit interaction information between the question and the picture. BAN+Counter further integrates Counter's counting module on the basis of BAN, effectively improving the performance of counting tasks and the overall performance.

4) MuRel, LCGN, and ReGAT outperform other comparison models, focusing on extracting information about entity relationships in images and playing a key role in joint feature learning. LCGN constructs a complete graph of the entity and implements implicit message passing between entities through problem guidance, while MuRel models the implicit relationship between detailed image regions through cross-modal attention, gradually refining the interaction between the picture and the problem. ReGAT explicitly extracts entity relationships in images and uses relational aware graph attention to realize accurate learning of joint features, so its performance is better than the other three models.

5) UFSCAN and MMMAN used more effective attention mechanisms. UFSCAN adopted a feature-based attention mechanism and obtained better results on counting questions by suppressing irrelevant features and emphasizing informative features. MMMAN proposed a multi-level mesh mutual attention, utilizing mutual attention to fully explore the information interaction between visual and language modalities and improve the model accuracy on Y/N questions.

6) Finally, these methods are inferior to our proposed method in accuracy. It is attributed to two points: (1) STCGN uses a syntax tree to model the question features, introduces syntactic structure features, and designs a tree hierarchical convolutional network to convolute the syntax tree structure, to fully extract the grammatical information of the question and improve its performance. (2) STCGN introduces a phrasing-aware message-passing module, which uses different phrase information in the question representation to guide the message passing between entities in multiple steps and extracts the context-aware entity features of the scene, which further promotes the performance of the model.

\subsection{Ablation Study}
\begin{figure}[ht]
    \centering
    \vspace{-4mm}
    \includegraphics[width=0.7\textwidth]{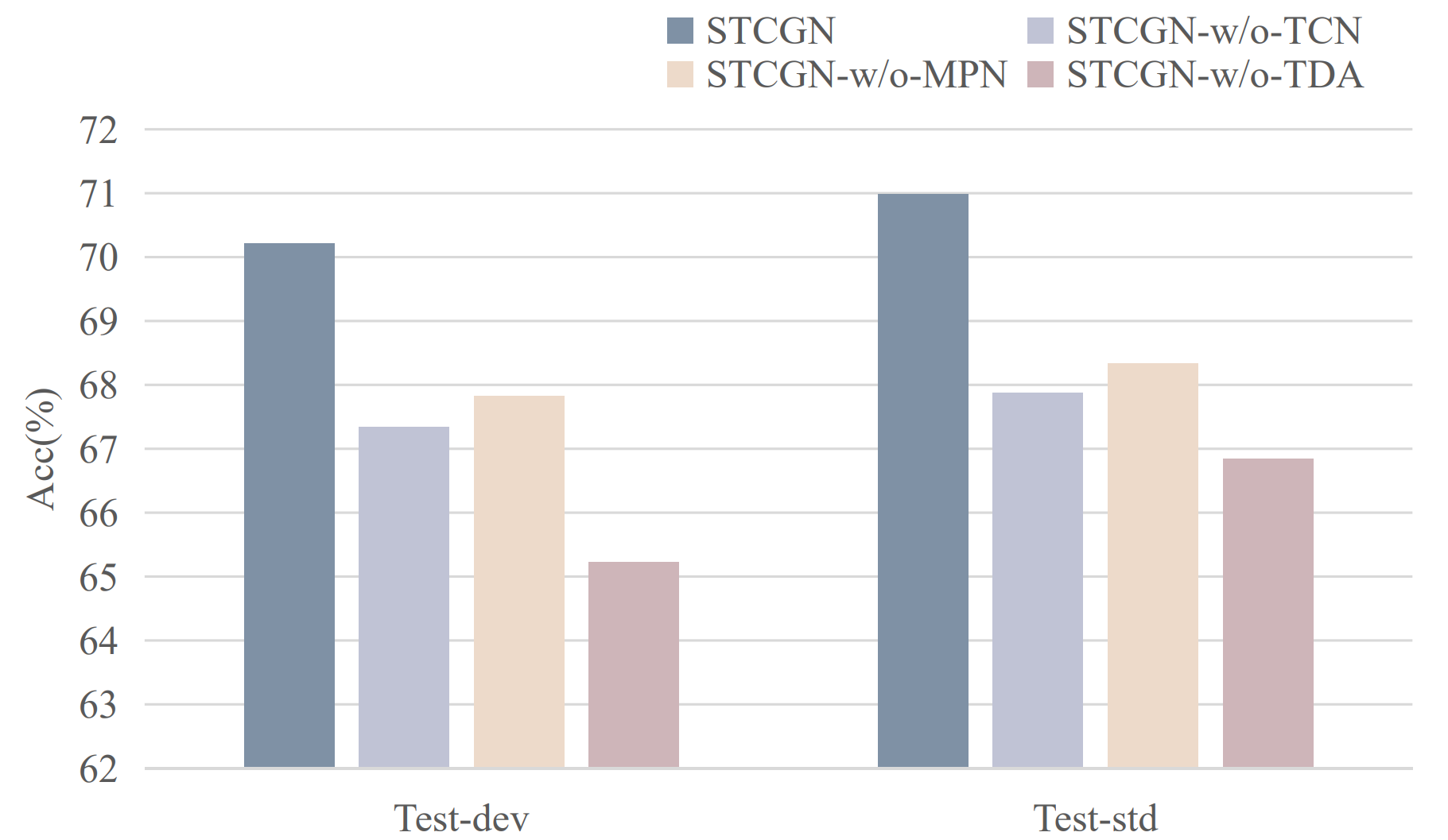}
    \vspace{-3mm}
    \caption{The overall accuracy of STCGN Variants.}
    \label{fig:ablation}
    \vspace{-7mm}
\end{figure}

Fig. \ref{fig:ablation} shows the performance of STCGN Variants. The results show that when any module is removed, the performance of STCGN on the Test-dev and Test-std subsets of VQA2.0 will decrease significantly. When the SHA module is lost, it has the greatest impact on the model, indicating that feature fusion is the module that has the greatest impact on the accuracy of visual question answering. Secondly, the influence of the TCN module on model performance is greater than that of the MPN module, which may be due to the fact that the tree convolution module is based on a syntax tree and plays an important role in extracting question features and guiding entities for message passing.

\subsection{Parameter Sensitivity}
\begin{figure}[ht]
    \vspace{-5mm}
    \centering
    \includegraphics[width=\textwidth]{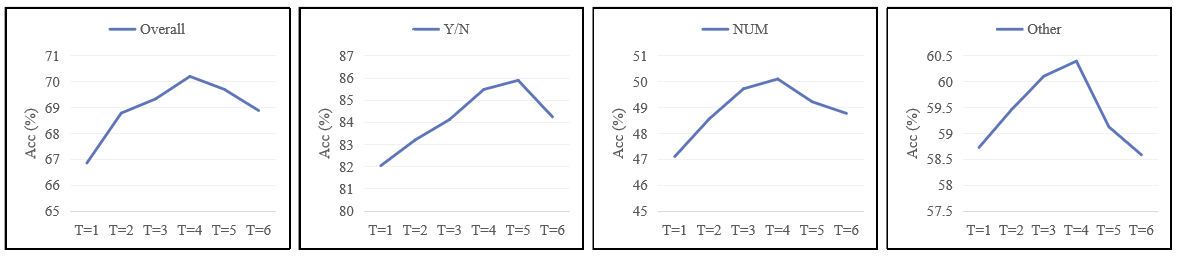}
    \vspace{-5mm}
    \caption{Parameter sensitivity of both iteration steps.}
    \label{fig:para}
    \vspace{-10mm}
\end{figure}

In this section, we analyze the effect of different iteration steps $T$. We fixed other parameters as the optimal parameters, gradually increased $T$, and obtained the curve of answering accuracy of different types of questions with the change of $T$, as shown in Fig. \ref{fig:para}. As $T$ increases, the rotation of messages between entities increases to incorporate more scene context information. The performance of all questions is progressively improved and optimally reached at $T=4$. As $T$ continues to increase, the performance of the model gradually decreases, since message passing leads to receive redundant information of the entity and reduce the accuracy of entity representation. The reason why the outliers appear is that the binary question has lower requirements for understanding questions and pictures than other questions, leading to better performance of the model in the binary questions but the inferior overall performance. Therefore, we chose $T=4$ as the final total number of messaging iteration steps.

\subsection{Attention Visualization}
\begin{figure}[!t]
    \vspace{-2mm}
    \centering
    \includegraphics[width=0.75\textwidth]{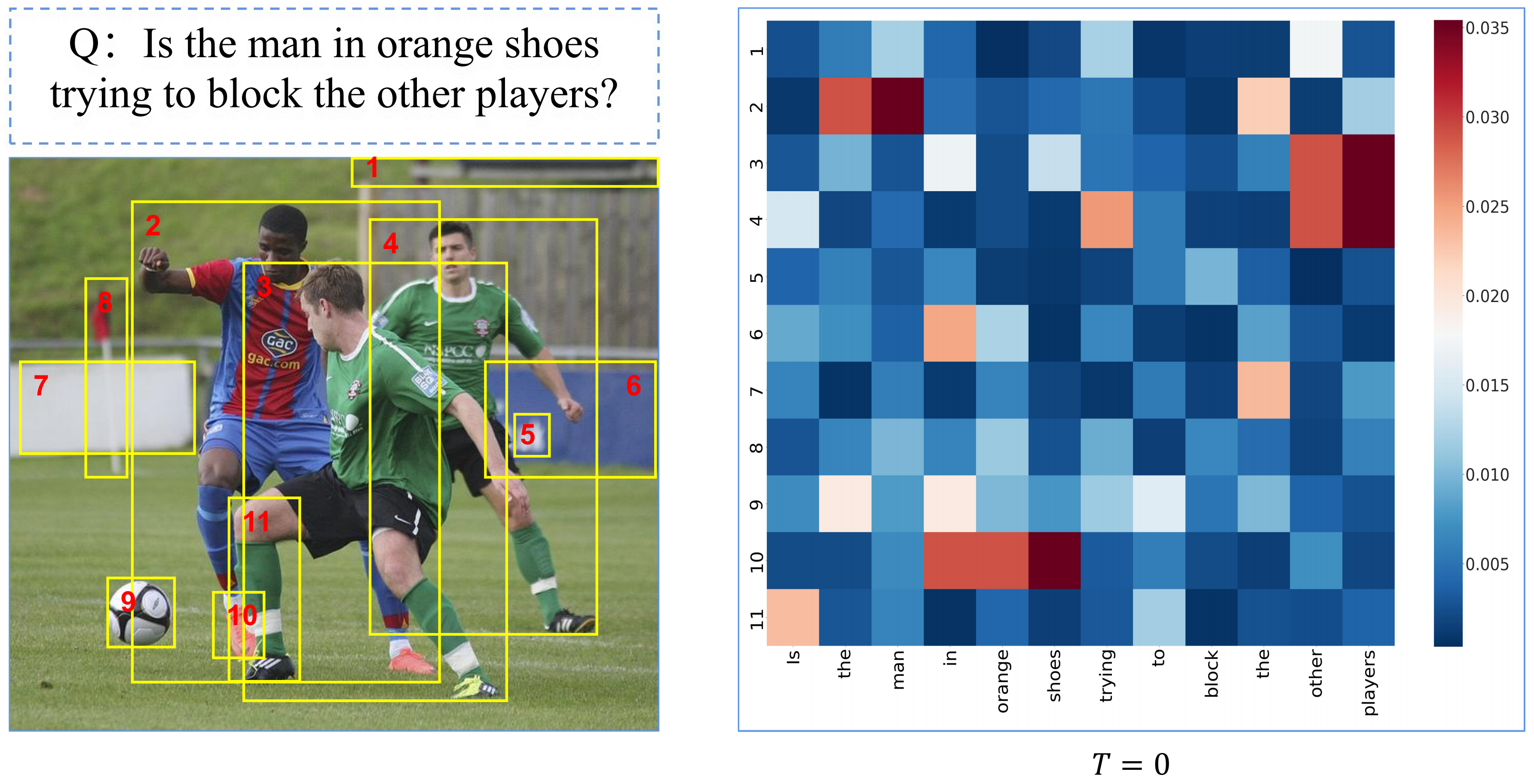}
    \includegraphics[width=0.75\textwidth]{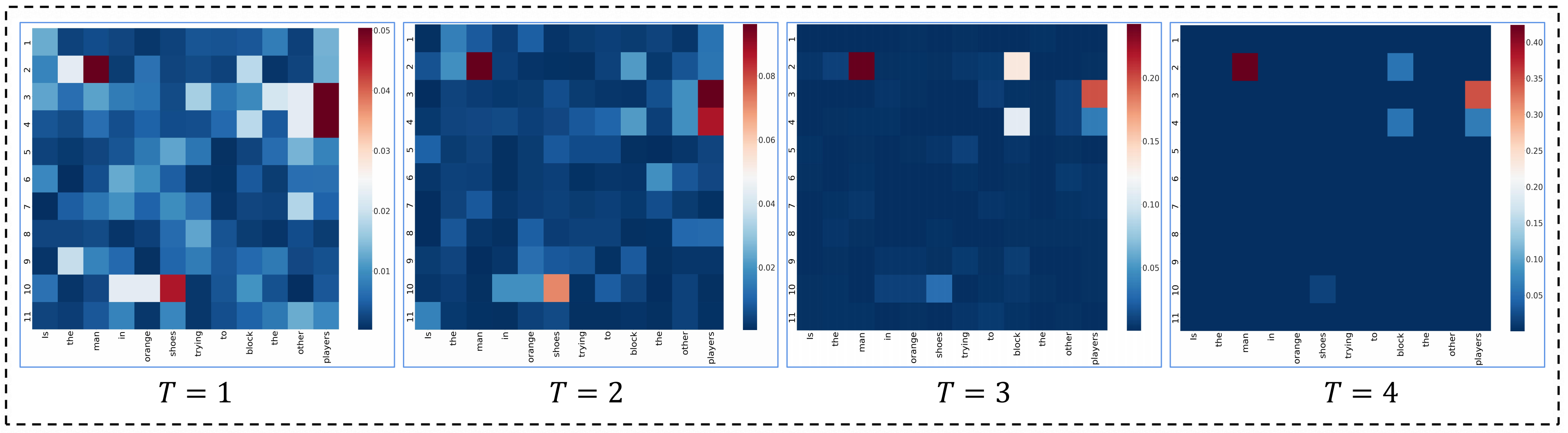}
    \vspace{-3mm}
    \caption{Visualization of attention score.}
    \label{fig:visual}
        \vspace{-5mm}
\end{figure}

To better illustrate the effectiveness of a phrase-based messaging mechanism, we experiment with visualizations in this section. We visualized the attention score between different entities in multiple iteration steps and different words of the question, as shown in Fig. \ref{fig:visual}. In the attention diagram, we can see: (1) Entity 2, Entity 4, and Entity 10 have significantly higher attention weights related to multiple phrases "the man", "in orange shoes" and "the other players" than other entity-word attention blocks. This suggests that the degree to which an entity is important in answering a question is closely related to multiple phrases. Syntax-aware phrase features in the messaging module provide guidance so that the VQA system can gradually understand entities that contribute more to the task. (2) The initial attention map can only initially locate important entities 2, 4, and 10, but their attention weights are not high. The initial visual attention map is very messy, and the contribution of each entity to the answer is not different. As the message passing iteration steps increase, the attention map becomes clearer and the attention weight of the key entity increases from 0.03 to 0.35. This is the result of non-critical entities passing information to critical entities, while also getting the entity representation of scene context awareness.

\section{Conclusion}
In this work, we propose a Syntax Tree Constrained Graph Network. We design a hierarchical tree convolutional network and extract phrase representation and question representation of syntactic structure perception from syntactic tree structure by combining text convolution with graph attention. At the same time, we also suggest a phrase-aware entity message-passing mechanism based on the observation of the data set. In multiple iteration steps, different instruction vectors are calculated using phrase features and question features to capture the scene context-aware entity features.

\bibliographystyle{splncs04}
\bibliography{camera-ready}

\begin{thebibliography}{10}
\providecommand{\url}[1]{\texttt{#1}}
\providecommand{\urlprefix}{URL }
\providecommand{\doi}[1]{https://doi.org/#1}

\bibitem{ref_butd}
Anderson, P., He, X., Buehler, C., Teney, D., Johnson, M., Gould, S., Zhang,
  L.: Bottom-up and top-down attention for image captioning and visual question
  answering. In: {CVPR}. pp. 6077--6086 (2018)

\bibitem{ref_simple}
Antol, S., Agrawal, A., Lu, J., Mitchell, M., Batra, D., Zitnick, C.L., Parikh,
  D.: {VQA:} visual question answering. In: {ICCV}. pp. 2425--2433 (2015)

\bibitem{ref_mutan}
Ben{-}Younes, H., Cad{\`{e}}ne, R., Cord, M., Thome, N.: {MUTAN:} multimodal
  tucker fusion for visual question answering. In: {ICCV}. pp. 2631--2639
  (2017)

\bibitem{ref_murel}
Cad{\`{e}}ne, R., Ben{-}Younes, H., Cord, M., Thome, N.: {MUREL:} multimodal
  relational reasoning for visual question answering. In: {CVPR}. pp.
  1989--1998 (2019)

\bibitem{guo2021bilinear}
Guo, D., Xu, C., Tao, D.: Bilinear graph networks for visual question
  answering. {IEEE} Trans. Neural Networks Learn. Syst.  (2021)

\bibitem{ref_lstm}
Hochreiter, S., Schmidhuber, J.: Long short-term memory. Neural Comput.
  \textbf{9}(8),  1735--1780 (1997)

\bibitem{ref_lcgn}
Hu, R., Rohrbach, A., Darrell, T., Saenko, K.: Language-conditioned graph
  networks for relational reasoning. In: {ICCV}. pp. 10293--10302 (2019)

\bibitem{ilievski2017generative}
Ilievski, I., Feng, J.: Generative attention model with adversarial
  self-learning for visual question answering. In: Thematic Workshops of ACM
  Multimedia 2017. pp. 415--423 (2017)

\bibitem{ref_ban}
Kim, J., Jun, J., Zhang, B.: Bilinear attention networks. In: NeurIPS. pp.
  1571--1581 (2018)

\bibitem{ref_dmn}
Kumar, A., Irsoy, O., Ondruska, P., Iyyer, M., Bradbury, J., Gulrajani, I.,
  Zhong, V., Paulus, R., Socher, R.: Ask me anything: Dynamic memory networks
  for natural language processing. In: {ICML}. vol.~48, pp. 1378--1387 (2016)

\bibitem{MMMAN}
Lei, Z., Zhang, G., Wu, L., Zhang, K., Liang, R.: A multi-level mesh mutual
  attention model for visual question answering. Data Sci. Eng.  \textbf{7}(4),
   339--353 (2022)

\bibitem{ref_regat}
Li, L., Gan, Z., Cheng, Y., Liu, J.: Relation-aware graph attention network for
  visual question answering. In: {ICCV}. pp. 10312--10321 (2019)

\bibitem{ref_coco}
Lin, T., Maire, M., Belongie, S.J., Hays, J., Perona, P., Ramanan, D.,
  Doll{\'{a}}r, P., Zitnick, C.L.: Microsoft {COCO:} common objects in context.
  In: {ECCV} {(5)}. vol.~8693, pp. 740--755 (2014)

\bibitem{nguyen2018improved}
Nguyen, D.K., Okatani, T.: Improved fusion of visual and language
  representations by dense symmetric co-attention for visual question
  answering. In: {CVPR}. pp. 6087--6096 (2018)

\bibitem{ref_glove}
Pennington, J., Socher, R., Manning, C.D.: Glove: Global vectors for word
  representation. In: {EMNLP}. pp. 1532--1543 (2014)

\bibitem{ren2015exploring}
Ren, M., Kiros, R., Zemel, R.: Exploring models and data for image question
  answering. {NIPS}  \textbf{28} (2015)

\bibitem{SimonyanZ14a}
Simonyan, K., Zisserman, A.: Very deep convolutional networks for large-scale
  image recognition. In: {ICLR} (2015)

\bibitem{ref_san}
Yang, Z., He, X., Gao, J., Deng, L., Smola, A.J.: Stacked attention networks
  for image question answering. In: {CVPR}. pp. 21--29 (2016)

\bibitem{YeH0LNL23}
Ye, T., Hu, L., Zhang, Q., Lai, Z.Y., Naseem, U., Liu, D.D.: Show me the best
  outfit for {A} certain scene: {A} scene-aware fashion recommender system. In:
  {WWW}. pp. 1172--1180 (2023)

\bibitem{yu2018beyond}
Yu, Z., Yu, J., Xiang, C., Fan, J., Tao, D.: Beyond bilinear: Generalized
  multimodal factorized high-order pooling for visual question answering. IEEE
  transactions on neural networks and learning systems  \textbf{29}(12),
  5947--5959 (2018)

\bibitem{ZhangCSN22}
Zhang, Q., Cao, L., Shi, C., Niu, Z.: Neural time-aware sequential
  recommendation by jointly modeling preference dynamics and explicit feature
  couplings. {IEEE} Trans. Neural Networks Learn. Syst.  \textbf{33}(10),
  5125--5137 (2022)

\bibitem{00200CSWL22}
Zhang, Q., Hu, L., Cao, L., Shi, C., Wang, S., Liu, D.D.: A probabilistic code
  balance constraint with compactness and informativeness enhancement for deep
  supervised hashing. In: {IJCAI}. pp. 1651--1657 (2022)

\bibitem{UFSCAN}
Zhang, S., Chen, M., Chen, J., Zou, F., Li, Y., Lu, P.: Multimodal feature-wise
  co-attention method for visual question answering. Inf. Fusion  \textbf{73},
  1--10 (2021)

\bibitem{zhang2015sensitivity}
Zhang, Y., Wallace, B.: A sensitivity analysis of (and practitioners' guide to)
  convolutional neural networks for sentence classification. arXiv preprint
  arXiv:1510.03820  (2015)

\end{thebibliography}

\clearpage
\appendix
\section{Experimental Settings}
\label{app:exp_settings}
\subsection{Datasets}
The dataset consists of real images from MSCOCO~\cite{ref_coco} with the same training, test, and validation set separation, with more than 80k images and 444k questions, 40k images and 214k questions, 80k images, and 448k questions. For each image, there are an average of at least three questions. The questions fall into three categories based on the type of response: yes/no, number, and others. Each pair collected 10 answers from human annotators, choosing the most frequent answer as the correct answer. This data set contains both open-ended and multiple-choice questions. In this paper, we focus on open-ended questions, taking answers that appear more than 9 times in the training set as candidate answers, and generating 3129 candidate answers.
\subsection{Baselines}
We use the following state-of-the-art methods as baselines to evaluate the performance of our STCGN: (1) LSTM-VGG~\cite{ref_simple} uses LSTM and VGG to extract text and image features respectively, and realizes the feature fusion through the element cross product form. (2) SAN~\cite{ref_san} designs a stacked attention mechanism that computes the most relevant areas of the question step by step to deduce the final answer. (3) DMN~\cite{ref_dmn} constructs four network modules of input, question, episodic memory, and answer. Through the cyclic work of these modules, an iterative attention process is generated. (4) MUTAN~\cite{ref_mutan} proposed a tensor-based Tracker decomposition method, which uses low-rank matrix decomposition to solve the problem of the large number of parameters in traditional bilinear models. (5) BUTD~\cite{ref_butd} proposed a combination of bottom-up and top-down attention mechanisms. Through the top-down attention mechanism, the correlation between the question and each region is calculated, so as to obtain more accurate picture features related to the question. (6) BAN~\cite{ref_ban} considers the bilinear interaction between two input channels is considered and the visual and textual information is fused by calculating the bilinear attention distribution. (7) LCGN~\cite{ref_lcgn} designs a question-aware messaging mechanism and uses question features to guide the refinement of entity features through multiple iteration steps, and realizes the integration of entity features and context information. (8) Murel~\cite{ref_murel} enriches the interaction between the problem region and the image region by vector representation and a sequence of units composed of multiple MuRel units.
(9) ReGAT~\cite{ref_regat} extracted explicit relation and implicit relation in each image and constructed explicit relation graph and implicit relation graph. At the same time, the graph attention network of question awareness is used to integrate the information of different relation spaces. (10)UFSCAN~\cite{UFSCAN} proposed a multimodal feature-wise attention mechanism and modeled feature-wise co-attention and spatial co-attention between image and question modalities simultaneously. (11)MMMAN~\cite{MMMAN} proposes a multi-level mesh mutual attention model to utilize low-dimensional and high-dimensional question information at different levels, providing more feature information for modal interactions.

\subsection{Settings}
In the experiments, Adamx was selected as the optimizer, and the initial learning rate was set at 0.001. With the increase in learning rounds, the learning rate gradually increased from 0.001 to 0.004. When the model accuracy reached its peak, the learning rate began to decay. During training, the decay rate of the learning rate was set to 0.5, and the model training rounds were 30 rounds. Regarding text, the word representation dimension of the question was 300, and the part of the speech embedding dimension was 128. In the tree hierarchical convolution module, word-level convolution uses a convolution kernel with the size of 3*428 for text convolution operation and the maximum syntax subtree length set to 4, and phrase-level convolution uses multi-head attention mechanism with the number of 8. To extract question features, we use a bidirectional GRU network with a hidden layer dimension of 1024. For picture features, we set the dimension as 2048, each picture contains 10-100 visual entity areas, the total time step of message passing $T$ is set as 4, and the dimension of scene context-aware entity feature is set as 1024. In our experiment, the VQA score provided by the official VQA competition is used as the evaluation metrics, shown as follows:
\begin{align}
    acc(\textbf{ans}) = min(\frac{\#human \ provides \ \textbf{ans}}{3},1)
\end{align}

\end{document}